# seeBias: A Comprehensive Tool for Assessing and Visualizing AI Fairness

Yilin Ning[1,2], Yian Ma[1,2], Mingxuan Liu[1,2], Xin Li[1,2], Nan Liu[1,2,3,4*]

[1]Centre for Quantitative Medicine, Duke-NUS Medical School, Singapore, Singapore

[2]Duke-NUS AI + Medical Sciences Initiative, Duke-NUS Medical School, Singapore, Singapore

[3]Programme in Health Services and Systems Research, Duke-NUS Medical School, Singapore, Singapore

[4]NUS Artificial Intelligence Institute, National University of Singapore, Singapore, Singapore

*Correspondence: Nan Liu, Centre for Quantitative Medicine, Duke-NUS Medical School, 8 College Road, Singapore 169857, Singapore. Phone: +65 6601 6503. Email: liu.nan@duke-nus.edu.sg

**Abstract**

Fairness in artificial intelligence (AI) prediction models is increasingly emphasized to support responsible adoption in high-stakes domains such as health care and criminal justice. Guidelines and implementation frameworks highlight the importance of both predictive accuracy and equitable outcomes. However, current fairness toolkits often evaluate classification performance disparities in isolation, with limited attention to other critical aspects such as calibration. To address these gaps, we present *seeBias*, an R package for comprehensive evaluation of model fairness and predictive performance. *seeBias* offers an integrated evaluation across classification, calibration, and other performance domains, providing a more complete view of model behavior. It includes customizable visualizations to support transparent reporting and responsible AI implementation. Using public datasets from criminal justice and healthcare, we demonstrate how *seeBias* supports fairness evaluations, and uncovers disparities that conventional fairness metrics may overlook. The R package is available on GitHub, and a Python version is under development.

# INTRODUCTION

Artificial intelligence (AI) is increasingly subject to governance and regulation across multiple ethical dimensions to ensure responsible and trustworthy use. A key consideration is fairness (1), where AI models should not only demonstrate satisfactory performance but also avoid systematically disadvantaging any population in decision-making, such as performing suboptimal for individuals based on race, ethnicity, sex, or gender. Rigorous evaluation of fairness is crucial to prevent perpetuating social inequalities and discrimination while building public trust. This emphasis on fairness is reflected in guidelines and frameworks, including the TRIPOD+AI guideline for responsible development and validation of clinical risk prediction model (2), the AI-Based Clinical Decision Support (ABCDS) framework for local health system implementation (3), and the Coalition for Health AI (CHAI) Responsible AI Guide that advocate continuous monitoring of AI for health, particularly in relation to fairness (4). Similar guides and requirements exist for high-stakes areas beyond healthcare, including criminal justice and finance, further highlighting the need for comprehensive computational tools to assess and ensure fairness in AI models.

The growing emphasis on fairness in AI prediction models has driven the development of a wide range of fairness metrics, each designed to quantify fairness and bias from different perspectives. A common approach is group fairness, which evaluates bias by comparing model predictions or performance between subgroups defined by sensitive variables. For example, demographic parity requires no differences in predicted risk across subgroups, while equal opportunity and equalized odds demand equal true positive and false positive rates between groups. To support practical implementation, fairness assessment toolkits such as AI Fairness 360 (*AIF360*) by IBM (5) and *Fairlearn* by Microsoft (6) incorporate these widely used group fairness metrics. Both toolkits offer visualizations to facilitate interpretation of fairness metrics, although these visualizations may be somewhat basic. The more recent *fairmodels* package covers a wider range of performance aspects in group fairness assessment (7), such as accuracy and positive/negative predictive values, while also providing more advanced visualizations. Additionally, all three toolkits offer various methods to mitigate bias identified in prediction models, helping ensure fairer predictions.

However, existing toolkits face certain limitations. Fairness metrics are typically derived from standard performance metrics, such as differences or ratios between subgroups. While these metrics are useful for detecting disparities, they may not provide a complete picture of model behavior across subgroups. Toolkits such as *AIF360* and *Fairlearn* offer flexibility by allowing users to select fairness metrics and incorporate performance metrics into evaluations and visualizations. However, this flexibility may pose challenges for users who are less familiar with fairness evaluation, particularly when creating standardized and comprehensive fairness reports. *Fairmodels* addresses this by providing a unified function to compute and visualize a wide range of fairness metrics, yet it does not integrate actual performance metrics. Furthermore, *fairmodels* requires prediction models of certain classes as input, which can be restrictive in real-world applications.

Focusing predominantly on fairness metrics also limits current toolkits in other important aspects of group fairness evaluation. For instance, the ABCDS framework highlights the importance of consistent model performance across subgroups in terms of calibration, in addition to classification (3). However, calibration is not easily captured by simple metrics (2,8,9), making it more complex to quantify than classification-based fairness (10,11). As a result, calibration-based fairness assessments are not directly implemented in toolkits discussed above. Similarly, rank-based fairness, which assesses the consistency of predicted risk or score rankings across subgroups (12), is also underexplored by current toolkits. Broadening the focus to performance metrics can also facilitate the exploration of emerging metrics that better communicate model performance to domain experts. For example, epidemiologists assess screening effectiveness by translating risk measures into the number needed to screen (NNS), i.e., the number of individuals who must be screened to prevent one adverse outcome (13,14). A recent study extrapolated NNS from the positive predictive value (PPV) of a test (15), offering a more intuitive way to communicate AI model performance. However, developing fairness metrics based on such concepts remains challenging, as they do not readily integrate into existing fairness evaluation toolkits.

To address these gaps, we developed *seeBias*, a toolkit for comprehensive fairness evaluation and visualization. *seeBias* incorporates a broader range of metrics for assessing group fairness, informed by recent advancements in fairness research and guidelines. It covers conventional parity-based and performance-based fairness evaluations, as well as calibration-based and rank-based assessments, with equal focus on predictive performance and fairness of AI models. Additionally, *seeBias* incorporates emerging practice that leverage concepts such as NNS to facilitate context-specific interpretation of fairness assessments. *seeBias* generates clear and customizable visualizations to facilitate reporting in scientific and operational contexts. Using two fully reproducible case studies from criminal justice and healthcare, we demonstrate the usability of *seeBias* and provide detailed interpretations of the results, offering practical guidance for future real-world applications.

# IMPLEMENTATION

AI fairness is a rapidly evolving field, with ongoing efforts to develop metrics and assessment methods to quantify fairness from different perspectives. While there is no clear consensus on which metrics best define fairness, some are widely used and offer broadly interpretable insights, whereas others are less common and difficult to quantify but can provide valuable perspectives for practical applications. To address this, *seeBias* implements functions for calculating commonly used fairness metrics to support established AI fairness practice and provides visualizations for less commonly examined aspects, enabling a more comprehensive fairness assessment.

**Established group fairness metrics**

AI fairness can be assessed from various perspectives, with a common approach focusing on consistency in model behavior across groups defined by sensitive variables (e.g., sex or race), known as group fairness. For example, equal opportunity ensures that different groups have the same true positive rate (TPR), indicating consistency in identification of positive cases across groups (16). Equalized odds imposes a stricter criterion by additionally requiring equal false positive rates (FPR) across groups (16). These metrics are widely used and implemented by *AIF360*, *Fairlearn*, and *fairmodels*. *seeBias* also implements these metrics by comparing the TPR and FPR of each group against a user-defined reference group. Additionally, *seeBias* implements balanced error rate (BER) equality, where BER is the average of FPR and the false negative rate (FNR, calculated as 1-TPR), to ensure consistent overall error rates across groups. These fairness metrics facilitate a quick assessment of disparities between groups.

Existing toolkits also implement demographic parity (or statistical parity), which requires equal proportion of predicted positives between groups. However, this approach can be controversial depending on the context and sensitive variables involved, e.g., when inspecting disease risks by sex or race (1,16,17). Hence, *seeBias* does not directly report metrics for demographic parity, but provides indirect insights through visualizations (see below for details).

**Detailed visualization for performance-based fairness evaluation**

Inspired by the comprehensive assessment in *fairmodels*, *seeBias* visually evaluates a wider range of performance metrics in addition to TPR and FPR to evaluate consistency in performance, including group-specific accuracy, PPV, and negative prediction value (NPV). Specifically, let TP, TN, FP, FN denote true positive, true negative, false positive, and false negative predictions for a group, and let n=TP+TN+FP+FN denote the group size. These performance metrics mentioned are defined as follows:

$$\text{Accuracy} = (TP + TN) / n,$$

$$TPR = TP / (TP + FN),$$

$$FPR = FP / (FP + TN),$$

$$PPV = TP / (TP + FP),$$

$$NPV = TN / (TN + FN).$$

While *fairmodels* emphasizes comparison with the reference group by reporting metrics ratios and highlights large disparities when ratio falls outside the 0.8 to 1.25 range, *seeBias* instead visualizes actual performance metrics to support both performance and fairness evaluations. Like *fairmodels*, *seeBias* indicates the 0.8 to 1.25 range relative to the reference level for each performance metric to facilitate group comparisons and identify notable disparities. This range is based on the 80% rule (or the "four fifths rule") originated from U.S. federal employment law, which serves as a general reference here but should not be interpreted as a definitive indicator of disparity (18–20).

**Additional visual investigation of calibration-based and rank-based fairness**

The performance-based group fairness measures discussed above focus on model ability to identify positive and negative samples at a given prediction threshold. An immediate extension is a receiver operating characteristic (ROC) analysis to ensure satisfactory and comparable performance across all groups. As emphasized in the TRIPOD+AI reporting guidelines for AI model development and validation (2) and in the ABCDS evaluation frameworks for their real-world deployment (3), calibration offers a more granular assessment of model performance and is essential for ensuring fair AI applications. To address this need, *seeBias* assesses calibration visually at two levels: calibration-in-the-large that compares observed and predicted positive proportions for each group at the given prediction threshold, and calibration curves that provide a detailed view of whether predicted probabilities align with observed proportions across risk levels. To facilitate accurate comparisons, the calibration-in-the-large plot reports the 95% confidence intervals (CIs) for observed positive proportions to account for sampling uncertainties. For smooth calibration curves, observed positive proportions at different risk levels are estimated using logistic recalibration (8). When model predictions are presented as risk scores rather than probabilities, Platt scaling is applied to transform scores into probabilities for calibration purposes through a logistic regression of observed labels on the scores (21,22).

Another important aspect of group fairness involves ensuring that the relative ranking of scores or probabilities for positive and negative samples is independent of sensitive variables (1,12). For instance, patients with a disease should consistently receive higher risk scores than those without, regardless of group distinctions based on sensitive attributes. *seeBias* assesses this by visualizing the distribution of predicted risk scores or probabilities across outcome labels and groups using boxplots. As will be demonstrated in Case Study 1, this approach can reveal systematic under- or over-estimation of risk for specific groups that may not be evident from ROC or calibration analyses.

When interpreting performance-based group fairness assessments, a practical challenge is the lack of a clear consensus on when disparities in performance constitute evidence of bias. While statistical tests can be conducted to evaluate the significance of differences between groups, statistical significance may not always reflect real-life relevance (1,23). To address this, *seeBias* draws inspiration from the epidemiological concept of NNS, which quantifies the efficacy of a screening test using the average number of individuals screened to identify one new case. Specifically, we interpret $1/PPV=(TP + FP)/TP$ as the average number of positive predictions needed to identify one true positive, termed number needed for true positive (NNTP). Similarly, 1/NPV=(TN + FN)/TN is interpreted as the average number of negative predictions required to identify one true negative (number needed for true negative [NNTN]). By visualizing NNTP and NNTN across groups and at different prediction thresholds, we enable straightforward comparisons of PPV and NPV differences, aiding in more intuitive interpretations of model fairness.

*seeBias* **package**

Based on the fairness assessment and visualizations described above, we design the *seeBias* package with three main components: specifying the necessary data to compute relevant metrics, generating a summary table for conventional fairness evaluation, and creating detailed visualizations for comprehensive evaluation of model performance and fairness. These components are visually summarized in Figure 1 and further elaborated below.

To ensure convenient usage and broad applicability, *seeBias* requires minimal input when computing performance and fairness metrics. Essential inputs include sensitive variables with reference level specified (if different from default values), observed binary labels and the label for positive class (if not using 0/1 encoding), and model predictions. Multiple sensitive variables are easily supplied as a matrix or data frame for intersectional fairness analysis. To accommodate model predictions in the form of probabilities and risk scores, *seeBias* offers two functions: *evaluate_prediction_prob()* and *evaluate_prediction_score()*, both allowing performance evaluation at prediction threshold derived from ROC analysis or specified by the user. The package also provides a function for binary predictions (*evaluate_prediction_bin()*) for basic evaluation and visualization. These functions create a *seeBias* object output to streamline further analysis.

Visualizations of fairness evaluations described in the previous section are implemented by applying the *plot()* function to the *seeBias* object. This creates a main visualization for group-specific performance metrics, and additional visualizations for ROC analysis, calibration analysis, prediction distribution, and the number needed. This enables researchers to perform a comprehensive visual inspection of group fairness from various perspectives, while also providing the flexibility to selectively report the most relevant visualizations for a more focused and succinct discussion. All visualizations are created with *ggplot2* (24) and can be easily customized. Conventional fairness evaluations are implemented by applying the *summary()* function to the *seeBias* object. This function generates a fairness evaluation table using equal opportunity, equalized odds, and BER equality, which compare TPR, FPR, and BER between groups defined by sensitive variables. By default, differences in metrics from the reference level are reported, with the option to report ratios instead. Definitions of these fairness metrics are provided for clarity.

## RESULTS

In this section, we describe two case studies in criminal justice and healthcare settings to demonstrate the interpretation of *seeBias* output. The seeBias package, and R code to reproduce the case studies, are available at https://github.com/nliulab/seeBias.

**Case Study 1: Intersectional Bias in Recidivism Prediction**

In this case study, we evaluated intersectional bias in predicting two-year recidivism using the widely studied Correlational Offender Management Profiling for Alternative Sanctions (COMPAS) dataset, sourced from the *fairmodels* R package and also provided in the *seeBias* package. The dataset includes demographic, criminal history, and offense information for 6,172 individuals. We focused on a subset of 5,278 White and Black individuals and examined bias across race and sex. We predicted two-year recidivism using a logistic regression model based on prior offenses, current charge (misdemeanor vs others), and age (under 25 or over 45 years). We used all individuals for model training and evaluation, and present their characteristics in Supplementary Table S1. R code to completely reproduce this case study, including the additional formatting to figures, is available at https://github.com/nliulab/seeBias.

Table 1 presents group sizes, and differences in TPR, FPR, and BER to facilitate fairness assessment. Results indicate higher TPR and FPR for Black, especially males, than for White males, reflecting violations of equal opportunity (equal TPR) and equalized odds (equal TPR and equal FPR). This discrepancy suggests that Black, particularly males, are overclassified as high-risk, potentially leading to unjust outcomes like stricter monitoring or denial of parole (25).

Figures 2 and 3 complement the conventional fairness analysis above by providing a broad set of performance metrics to evaluate the suitability of the model for real-life application, for example in terms of classification accuracy, AUC, and calibration. These metrics also enable more in-depth fairness assessments for additional insights. Where feasible, 95% CIs are reported to account for estimation uncertainty when comparing between groups for fairness assessment. In Figure 2, we observed similar accuracy across groups, though PPV and NPV varied, with higher PPV for Black males and higher NPV for Black females relative to White males. In addition to the disparities in TPR and FPR between White and Black individuals shown in Table 1, Figure 2 further highlights a low TPR for White individuals (especially below 0.5 for White females) that should be improved in subsequent bias mitigation.

In further assessment in Figure 3, group disparities in model performance are not immediately apparent from the ROC curves (Figure 3A), despite slightly higher AUC for Black individuals than White individuals. The calibration curve (Figure 3B) and calibration-in-the-large plot (Figure 3C) indicated an overestimation of risk for Black females and some overfitting of risk for the other three groups, with predicted risks more extreme than observed risks. Specifically, for low-risk individuals (i.e., White males and females) the predicted risks were lower than observed levels, and for high-risk individuals (i.e., Black males) the opposite was observed. Additionally, Figure 3D shows that Black individuals without recidivism received similar predicted risk to White individuals of the same sex with recidivism, highlighting racial biases in model predictions. These help pinpoint the inadequacy of the current prediction model in correctly predicting reoffenders to mitigate bias in downstream processes.

We further explored the practical implications of disparities observed in Figure 2 by converting PPV and NPV values to NNTP (Figure 3E) and NNTN (Figure 3F) across various prediction thresholds. The more pronounced disparities in PPV compared to NPV resulted in a maximum between-group difference of approximately one individual for NNTP and less than one for NNTN. Domain experts may evaluate whether these differences have practical significance and necessitate further investigation or action.

**Case Study 2: Clinical Fairness Evaluation in ROSC Prediction**

In this case study, we evaluated racial bias in predicting return of spontaneous circulation (ROSC) among out-of-hospital cardiac arrest (OHCA) patients using data from the Resuscitation Outcomes Consortium Cardiac Epidemiologic Registry (Version 3) (26). This study was exempted from National University of Singapore Institutional Review Board review. The final cohort included 58,648 patients aged 18 years or above, who were transported to hospital by emergency medical services (EMS), received resuscitation, and had complete data on ROSC status and 12 candidate variables. Cohort selection details are provided in Supplementary Appendix B, and sample characteristics are summarized in Supplementary Table S2.

To illustrate the assessment of bias, we developed a scoring model for ROSC prediction using the AutoScore framework (27,28), focusing on interpretability and simplicity for clinical use. The final scoring model included six predictors: initial rhythm, EMS response time, age, witness status, use of epinephrine, and use of any medication (see Supplementary Appendix B for model development). Racial bias was assessed based on predicted scores in the test set (n=11,730), using White patients as the reference group. R code for model development and bias assessment is available at https://github.com/nliulab/seeBias.

The small differences in TPR, FPR and BER between groups (Table 2) did not provide strong evidence of racial disparities. However, a closer examination of model performance in Figure 4 revealed significant racial disparities in PPV and NPV, with Black patients showing notably lower PPV and Asian patients lower NPV. Minor differences in TPR and accuracy were also observed. Further analysis indicated a lower AUC for Asian, Black, and Hispanic patients compared to White and Others (Figure 5A). Only White and Others had AUC close to 0.7 (0.704, 95% CI: 0.692-0.716 for Others; 0.691, 95% CI: 0.671-0.711 for White). Although the 95% CI for Asian and Hispanic patients included 0.7, the wide intervals highlight substantial uncertainty due to small sample sizes (Table 2). Calibration curves were generally aligned with the diagonal for White and Others, but there was pronounced underestimation of ROSC probability for Asian patients and some overestimation for Hispanic and Black patients (Figure 5B). These miscalibrations may arise from the similar distribution of predicted probabilities across groups (Figure 5D), which failed to reflect differences in observed positive rates, resulting in pronounced underestimation for Asian patients and overestimation for Black and Others (Figure 5C).

Given the imbalanced representation of racial groups in this cohort (Table 2), and the modest accuracy (Figure 4) and AUC (Figure 5A), incorporating additional samples from underrepresented groups may improve the performance of the prediction model for clinical applications. Translating PPV and NPV into number of patients (Figures 5E and 5F) further emphasized the disparities between groups. Evaluated at the prediction threshold of 51.5 from the ROC analysis (Figure 4), the maximum between-group difference was approximately one individual for NNTP and less than one individual for NNTN. These results offer additional data-driven insights into the suitability of the model for fair ROSC prediction across racial groups.

## DISCUSSION

Bias in AI models has been a significant concern in high-stakes applications such as criminal justice and healthcare (1,29–31). Current efforts to evaluate and control algorithmic bias often adopt a fairness-centric approach, separate from standard performance evaluations. Software implementations of these approaches further contribute to their widespread adoption. However, such fairness-driven strategies may be inadequate for developing reliable and equitable AI models, as fairness can sometimes be achieved at the cost of reduced predictive performance, and the mechanisms underlying these trade-offs are not yet fully understood (12,32). Additionally, most fairness metrics rely on group comparisons of model classification, while other critical aspects, such as calibration or risk ranking across groups, are more difficult to incorporate and often overlooked. To address these limitations, we propose a comprehensive performance-centric group fairness evaluation and that includes calibration- and rank-based assessments, using visualizations support more granular analysis that are challenging to capture with single metrics. This is consistent with recent recommendations that fairness assessment of AI prediction models should go beyond single metrics to better characterize trade-offs and potential harms (30). We implement this approach in the *seeBias* package, providing detailed fairness evaluations and high-quality figures to facilitate fairness assessment in research and applications.

Similar to existing software toolkits, *seeBias* provides a main visualization for common group fairness assessments based on classification performance metrics (e.g., equal opportunity, equalized odds, and BER equality). It displays group-specific performance metrics with 95% CIs to account for sampling errors and separately reports fairness metrics in a table to facilitate reporting. In addition to identifying bias, this direct assessment of model performance highlights areas for improvements in bias mitigation, such as the inadequate TPR for White individuals in Case Study 1 on two-year recidivism prediction. These assessments can also reveal disparities in group performance that are not captured by conventional fairness metrics, as demonstrated in Case Study 2 on ROSC prediction.

Beyond classification-based fairness evaluations, *seeBias* provides visualizations to enable a more in-depth investigation of identified biases. For example, although the model in Case Study 1 performed consistently in predictive discrimination within race and sex groups, calibration curves revealed overprediction, where model predictions were more extreme than observed outcomes. Additionally, the distribution of predictions showed a notably higher predicted risk for Black individuals compared to White individuals. These findings reflect model inadequacy, as it learned historical biases present in the data, which would perpetuate bias if deployed in practice. In Case Study 2, ROC and calibration curves revealed more complex algorithm biases, including varying degrees of miscalibration and predictive discrimination across racial groups. As shown in the calibration-in-the-large plot, the model demonstrated demographic parity by not explicitly incorporating race in predictions but failed to account for observed racial differences in ROSC rates in the data. This raises important questions of whether the model should be adjusted to better capture these differences or whether the observed disparities reflect underlying biases that the model should help mitigate. Addressing this issue requires additional clinical input.

In developing *seeBias*, we emphasize the presentation of fairness investigations that more effectively align with context-specific interpretations. For example, we excluded demographic parity from the main visualization for group fairness evaluation and instead include it as part of the calibration-in-the-large plot. This is in line with the understanding that it is not always reasonable to expect similar risk levels across groups, such as when biological differences between males and females affect disease mechanisms (1). In such cases, it may be more meaningful to analyze prediction differences relative to observed outcomes and interpret them within the specific context, rather than automatically treating them as evidence of algorithm bias. Additionally, we aim to provide alternative representations of fairness measures to help assess the clinical significance of observed group differences, as statistical significance along often fails to capture the practical impact of these disparities. As an initial effort, we translate PPV and NPV to the number of individuals needed to obtain each true positive or true negative, inspired by the adaptation of NNS beyond epidemiological analysis of diagnostic test efficacy (15). We believe further exploration of such translations can help evaluate the real-world significance of group disparities, improving the interpretability of fairness studies for domain experts. This approach also reduces unnecessary attempts to mitigate bias, as poorly informed mitigation efforts can inadvertently harm both model fairness and performance (32).

Apart from enabling improved interpretations, we make *seeBias* user-friendly by simplifying its usage. The *seeBias* package is a lightweight tool dedicated to comprehensive fairness evaluation, guided by emerging AI guidelines and governance frameworks, with minimal input requirements for AI models to streamline its application. As demonstrated in our online guidebook (see https://github.com/nliulab/seeBias), users can easily adjust the style and arrangement of figures generated by *seeBias* using simple commands for better display. While we encourage comprehensive fairness evaluation and reporting, users have the flexibility to report only the subset of results most relevant to their specific aims and needs. This simplicity and flexibility make *seeBias* highly accessible to practitioners with varying levels of programming experience, facilitating the growing demand for fairness evaluations.

However, the simplicity of *seeBias* is achieved by excluding built-in pipelines for bias mitigation, an important feature in existing tools such as *AIF360*, *Fairlearn*, and *fairmodels*. Visualizations from these tools can accommodate multiple models to assess the effectiveness of bias mitigation or compare various bias mitigation methods, which is challenging for *seeBias* visualizations. Consequently, when bias mitigation is required, *seeBias* can complement outputs from these toolkits by offering detailed evaluations and comparisons of model performance. Additionally, by comprehensively evaluating model performance from multiple perspectives, *seeBias* facilitates deeper investigations into the root cause of identified bias and promotes a more informed selection of mitigation methods, moving beyond a single focus on changes in fairness metrics. Another limitation of seeBias is the difficulty in handling comparisons across a large number of groups. This can arise either from detailed categorization of a single sensitive variable, or from studying intersectional fairness across multiple sensitive variables. The current implementation is optimized for up to seven groups, and future development aims to address this limitation. Currently implemented in R, we are also developing a Python version and a graphical user interface to expand its accessibility and usability.

The importance of fairness evaluation in AI models has been increasingly recognized, as evidenced by the growing availability of software tools. However, the scope of these evaluations and the interpretation of fairness metrics continue to evolve with advancing understanding, alongside emerging guidelines and requirements. *seeBias* incorporates these developments by enabling comprehensive fairness assessments that extend beyond conventional fairness metrics, emphasizing actual model performance to provide a more complete picture of model reliability for deployment. Designed with lightweight commands, *seeBias* facilitates broad usage across diverse research areas and application settings. By providing these capabilities, we aim to support the development and deployment of fair AI models, contributing to a more equitable and responsible integration of AI in critical decision-making process.

**Data and code availability**

The *seeBias* package and R code to reproduce the two case studies are publicly available on GitHub at https://github.com/nliulab/seeBias. The COMPAS data is provided within the *seeBias* package. The Resuscitation Outcomes Consortium Cardiac Epidemiologic Registry (Version 3) is publicly available subject to ethical approval and compliance with the data use agreement.

**Table 1.** Intersectional fairness assessment of two-year recidivism between groups defined by race and sex, calculated by differences from the reference group.

| Group | Sample size | TPR difference | FPR difference | BER difference |
|---|---|---|---|---|
| White & Male | 1,621 | Reference | Reference | Reference |
| Black & Female | 549 | 0.173 | 0.13 | -0.021 |
| Black & Male | 2,626 | 0.242 | 0.194 | -0.024 |
| White & Female | 482 | -0.096 | -0.026 | 0.035 |

**Table 2.** Fairness assessment of ROSC prediction between groups defined by race, calculated by differences from the reference group.

| Group | Sample size | TPR difference | FPR difference | BER difference |
|---|---|---|---|---|
| White | 13,332 | Reference | Reference | Reference |
| Asian | 839 | -0.037 | 0.029 | 0.033 |
| Black | 5,986 | -0.040 | 0.042 | 0.041 |
| Hispanic | 1,521 | -0.019 | -0.012 | 0.004 |
| Others | 36,970 | 0.037 | 0.008 | -0.015 |

**Figure 1.** Main components of the *seeBias* package.

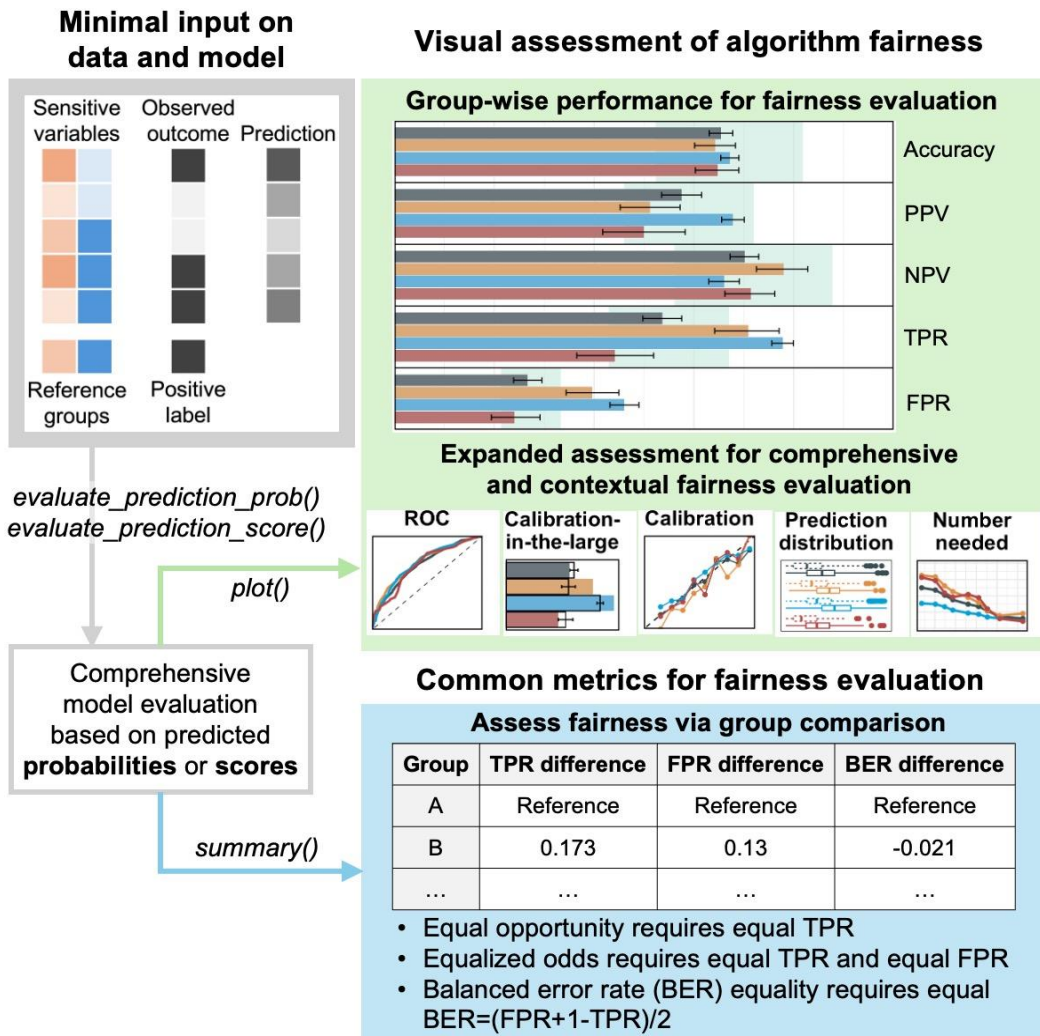

**Figure 2.** Intersectional fairness assessment of two-year recidivism prediction by comparing classification performance between groups defined by race and sex.

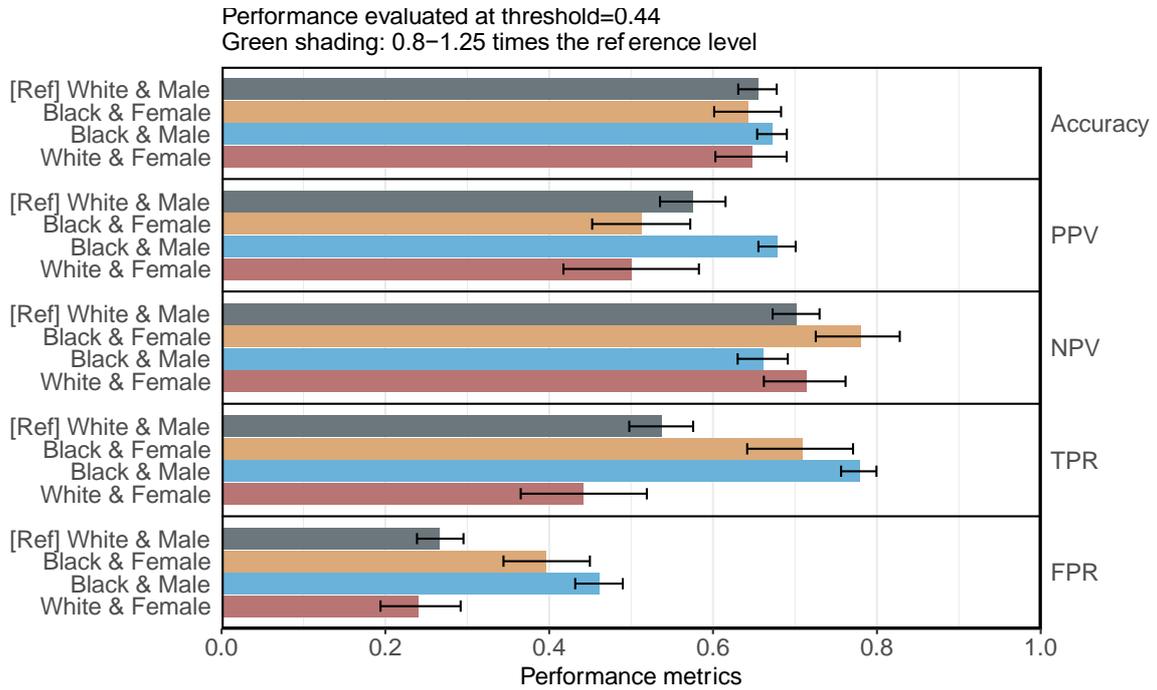

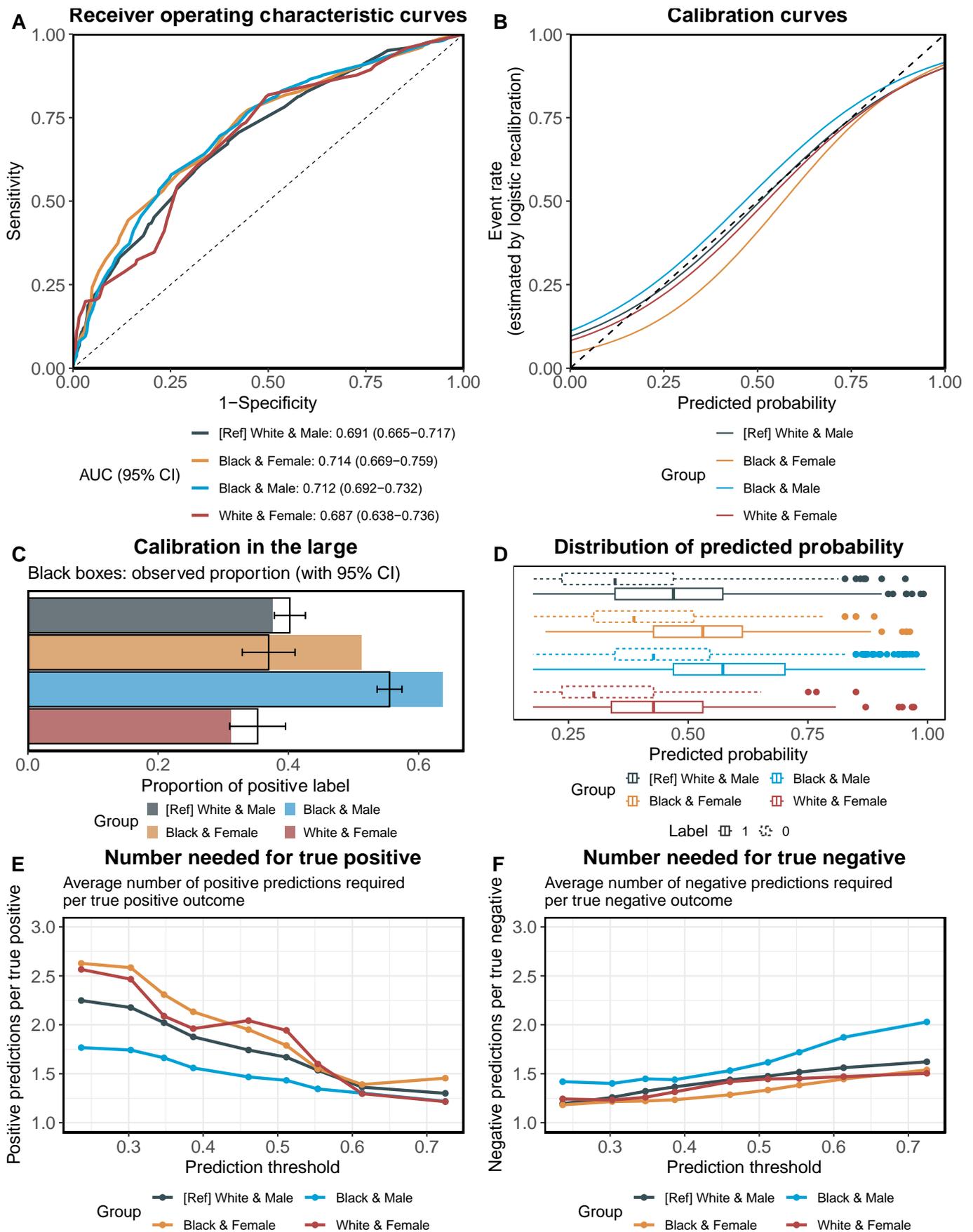

**Figure 3.** Additional assessment of intersectional fairness of two-year recidivism prediction.
22

**Figure 4.** Fairness assessment of return of spontaneous circulation prediction by comparing classification performance between groups defined by race.

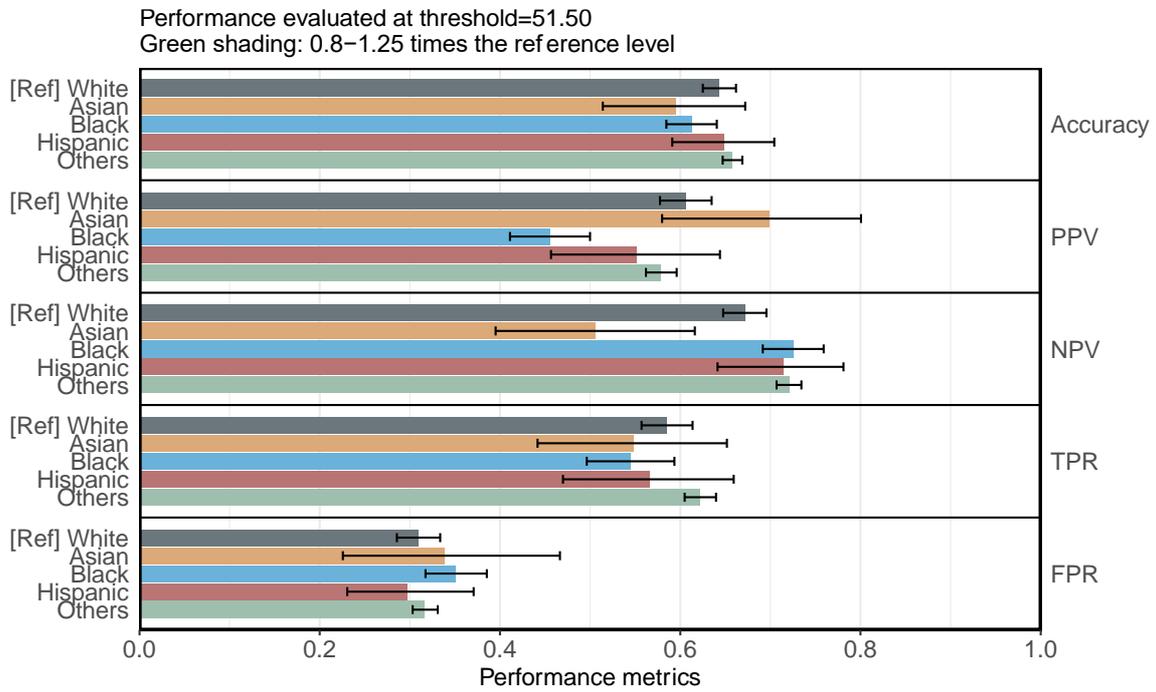

**Figure 5.** Additional assessment of fairness of return of spontaneous circulation prediction.

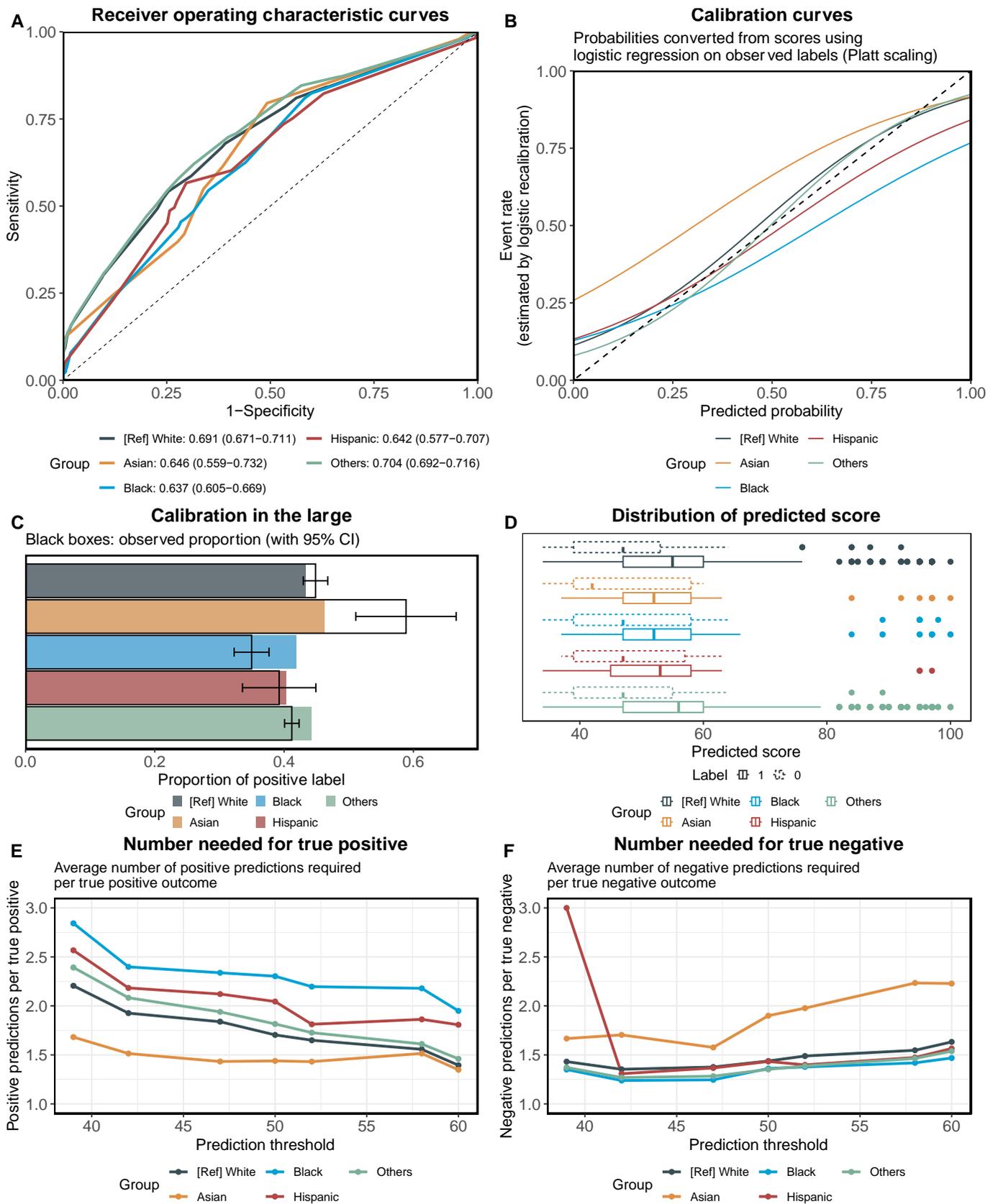

# Supplementary Appendix

## A. Case Study 1: Intersectional Bias in Criminal Justice

**Supplementary Table S1.** Characteristics of individuals analyzed in the prediction of two-year recidivism.

|  | Overall (n=5,278) | Did not have two-year recidivism (n=2,795, 53.0%) | Had two-year recidivism (n=2,483, 47.0%) |
| --- | --- | --- | --- |
| Prior offences: mean (SD) | 3.46 (4.88) | 2.14 (3.47) | 4.94 (5.73) |
| Age > 45: n (%) | 1096 (20.8) | 734 (26.3) | 362 ( 14.6) |
| Age < 25: n (%) | 1156 (21.9) | 496 (17.7) | 660 ( 26.6) |
| Misdemeanor: n (%) | 1838 (34.8) | 1113 (39.8) | 725 ( 29.2) |
| White: n (%) | 2103 (39.8) | 1281 (45.8) | 822 ( 33.1) |
| Male: n (%) | 4247 (80.5) | 2137 (76.5) | 2110 ( 85.0) |

SD: standard deviation.

## B. Case Study 2: Clinical Fairness Evaluation in ROSC Prediction

### Cohort Selection

Data for this case study was obtained from the Resuscitation Outcomes Consortium Cardiac Epidemiologic Registry (Version 3), a prospective, population-based registry of out-of-hospital cardiac arrests (OHCA) across eight regions in the United States and three in Canada (1). The binary outcome was defined as return of spontaneous circulation (ROSC), either before hospital arrival or upon arrival at the emergency department, with 1 indicating ROSC and 0 otherwise.

We extracted 12 candidate variables relevant to ROSC prediction, including patient demographics (age, sex), OHCA characteristics (cause of arrest, witness status, location of arrest, initial rhythm), emergency medical services (EMS) response time, and prehospital interventions (bystander cardiopulmonary resuscitation [CPR], mechanic CPR, use of epinephrine, use of amiodarone, and use of any medication). We also extracted race information for fairness assessment. Patients with multiple or unknown race were recoded as "Others" to simplify analysis.

The final cohort included 58,648 patients aged 18 years or above, who were transported to hospital by emergency medical services (EMS), received resuscitation, had a recorded sex (male or female), and had complete data on ROSC status and all variables of interest. Cohort characteristics are summarized in Supplementary Table S2.

**Supplementary Table S2.** Characteristics of individuals analyzed in the prediction of return of spontaneous circulation (ROSC) after out-of-hospital cardiac arrest.

|  | Overall (n=58,648) | Did not have ROSC (n=34,264, 58.4%) | Had ROSC (n=24,384, 41.6%) |
|---|---|---|---|
| Age: mean (SD) | 65.57 (16.70) | 65.57 (16.90) | 65.56 (16.41) |
| Male: n (%) | 37500 (63.9) | 22066 (64.4) | 15434 (63.3) |
| Race: n (%) | | | |
|   Asian | 839 (1.4) | 347 (1.0) | 492 (2.0) |
|   Black | 5986 (10.2) | 4012 (11.7) | 1974 (8.1) |
|   Hispanic | 1521 (2.6) | 912 (2.7) | 609 (2.5) |
|   White | 13332 (22.7) | 7301 (21.3) | 6031 (24.7) |
|   Others | 36970 (63.0) | 21692 (63.3) | 15278 (62.7) |
| Cardiac cause: n (%) | 54709 (93.3) | 32115 (93.7) | 22594 (92.7) |
| Witnessed arrest: n (%) | 30690 (52.3) | 14644 (42.7) | 16046 (65.8) |
| Location: n (%) | | | |
|   Healthcare | 4989 (8.5) | 3150 (9.2) | 1839 (7.5) |
|   Private | 44837 (76.5) | 26947 (78.6) | 17890 (73.4) |
|   Public | 8822 (15.0) | 4167 (12.2) | 4655 (19.1) |
| Initial rhythm: n (%) | | | |
|   Asystole | 26871 (45.8) | 19659 (57.4) | 7212 (29.6) |
|   PEA | 13960 (23.8) | 7049 (20.6) | 6911 (28.3) |
|   Unknown | 4519 (7.7) | 2436 (7.1) | 2083 (8.5) |
|   VF/VT | 13298 (22.7) | 5120 (14.9) | 8178 (33.5) |
| Response time: mean (SD) | 6.16 (3.44) | 6.22 (3.49) | 6.09 (3.36) |
| BCPR by: n (%) | | | |
|   Laypeople | 17768 (30.3) | 9749 (28.5) | 8019 (32.9) |
|   Other | 7368 (12.6) | 4462 (13.0) | 2906 (11.9) |
|   Unknown | 33512 (57.1) | 20053 (58.5) | 13459 (55.2) |
| Mechanic CPR: n (%) | 2395 ( 4.1) | 1620 ( 4.7) | 775 ( 3.2) |
| Used any drug: n (%) | 50729 (86.5) | 29977 (87.5) | 20752 ( 85.1) |
| Used epinephrine: n (%) | 48813 (83.2) | 29891 (87.2) | 18922 ( 77.6) |
| Used amiodarone: n (%) | 2872 ( 4.9) | 1376 ( 4.0) | 1496 ( 6.1) |

CPR: Cardiopulmonary resuscitation. BCPR: bystander CPR. PEA: Pulseless Electrical Activity. SD: standard deviation. VF/VT: Ventricular Fibrillation/Ventricular Tachycardia.

**Model Development Using AutoScore**

AutoScore is an automated machine learning framework designed to develop interpretable point-based scoring models for clinical risk prediction (2). It streamlines variable selection, score generation, and model evaluation through a series of systematic steps. First, candidate variables are ranked by predictive importance using machine learning algorithms, such as random forests. Then, subsets of variables are iteratively evaluated to achieve an optimal balance between model simplicity and predictive performance. Variables are categorized into clinically meaningful groups, and a scoring system is constructed by assigning point values to each category. The model is fine-tuned using a validation set to enhance clinical interpretability and performance. Finally, predictive performance is assessed on an independent test set.

In this case study, the final cohort was randomly split into training (70%, n=41,054), validation (10%, n=5,864), and testing (20%, n=11,730) sets, stratified by ROSC status to preserve outcome distribution. The 12 candidate variables were ranked using a random forest with 100 trees to generate a parsimony plot (see Supplementary Figure S1) to guide variable selection. The final scoring model was developed using the top six variables, as predictive performance showed minimal improvements with the inclusion of additional variables.

**Supplementary Figure S1.** AutoScore parsimony plot for the prediction of return of spontaneous circulation.

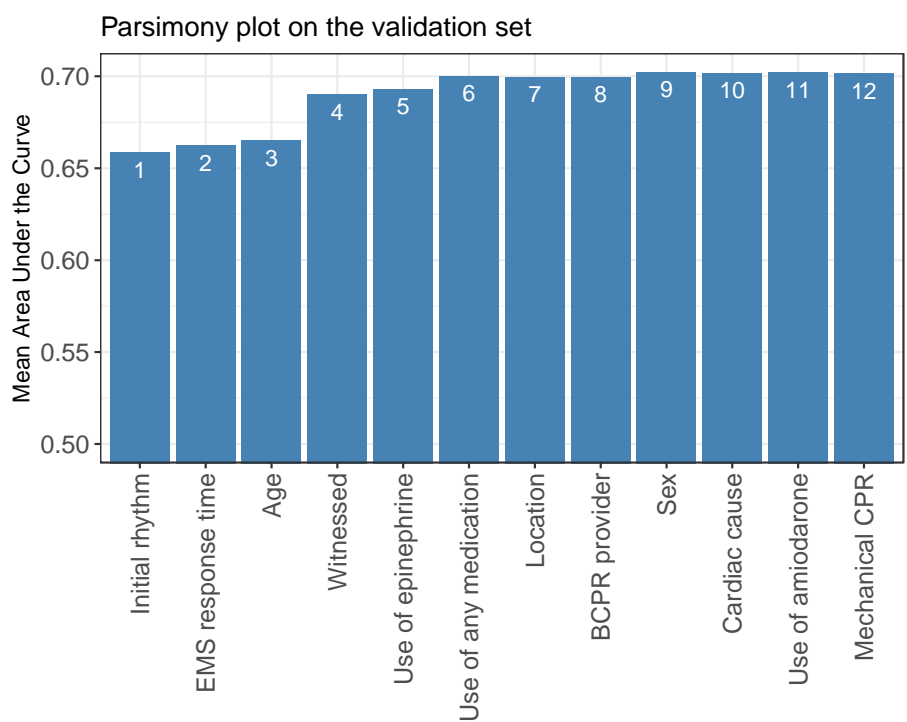

By default, continuous variables, i.e., response time and age in this example, were categorized based on percentiles from the training set. To improve clinical interpretability, we fine-tuned the cutoff values to 4, 8, and 12 minutes for response time, and 35 years for age. The scoring table of the final model is presented in Supplementary Table S3, which was be used to assign scores to patients in the test set for fairness evaluation.

**Supplementary Table S3.** Scoring table for predicting return of spontaneous circulation.

| Variable | Interval | Point |
|---|---|---|
| Initial rhythm | Asystole | 0 |
| | PEA | 11 |
| | Unknown | 8 |
| | VF/VT | 13 |
| Response time | <4 | 5 |
| | [4,8) | 5 |
| | [8,12) | 3 |
| | >=12 | 0 |
| Age | <35 | 3 |
| | >=35 | 0 |
| Witnessed | No | 0 |
| | Yes | 8 |
| Use of epinephrine | 1 | 0 |
| | 0 | 37 |
| Use of any drug | 1 | 34 |
| | 0 | 0 |

PEA: Pulseless Electrical Activity. VF/VT: Ventricular Fibrillation/Ventricular Tachycardia.

**Supplementary References:**